\documentclass[sigconf, nonacm]{acmart}

\usepackage{microtype}
\usepackage[skip=2pt]{caption}
\usepackage{algorithm}[H] 
\usepackage[noend]{algpseudocode} 
\usepackage{enumitem} 
\AtBeginDocument{%
  \providecommand\BibTeX{{%
    \normalfont B\kern-0.5em{\scshape i\kern-0.25em b}\kern-0.8em\TeX}}}

\begin{document}

\title{Learning to Walk Autonomously via Reset-Free Quality-Diversity}



\author{Bryan Lim}
\authornote{Both authors contributed equally to this research.}
\affiliation{
  \institution{Imperial College London}
  \country{United Kingdom}
}
\email{bryan.lim16@imperial.ac.uk}
\author{Alexander Reichenbach}
\authornotemark[1]
\affiliation{%
  \institution{Imperial College London}
  \country{United Kingdom}
  }
\email{alexander.reichenbach20@imperial.ac.uk}

\author{Antoine Cully}
\affiliation{%
  \institution{Imperial College London}
  \country{United Kingdom}
}
\email{a.cully@imperial.ac.uk}

\renewcommand{\shortauthors}{B. Lim, A. Reichenbach and A. Cully}

\begin{abstract}
Quality-Diversity (QD) algorithms can discover large and complex behavioural repertoires consisting of both diverse and high-performing skills.
However, the generation of behavioural repertoires has mainly been limited to simulation environments instead of real-world learning. 
This is because existing QD algorithms need large numbers of evaluations as well as episodic resets, which require manual human supervision and interventions. 
This paper proposes Reset-Free Quality-Diversity optimization (RF-QD) as a step towards autonomous learning for robotics in open-ended environments.
We build on Dynamics-Aware Quality-Diversity (DA-QD) and introduce a behaviour selection policy that leverages the diversity of the imagined repertoire and environmental information to intelligently select of behaviours that can act as automatic resets. 
We demonstrate this through a task of learning to walk within defined training zones with obstacles.  
Our experiments show that we can learn full repertoires of legged locomotion controllers autonomously without manual resets with high sample efficiency in spite of harsh safety constraints.
Finally, using an ablation of different target objectives, we show that it is important for RF-QD to have diverse types solutions available for the behaviour selection policy over solutions optimised with a specific objective. 
Videos and code available at \url{https://sites.google.com/view/rf-qd}.
\end{abstract}



\keywords{Quality-Diversity, Reset-free, Safety-aware Learning, Robotics}



\maketitle
\newcommand{\algonamefull}{Dynamics-Aware Quality-Diversity}
\newcommand{\algoname}{DA-QD}

\newcommand{\dynamicsmodel}{\widetilde{p}_{\vec \theta}}
\newcommand{\replaybuffer}{\mathcal{B}}

\newcommand{\archivereal}{\mathcal{A}}
\newcommand{\archivesynthetic}{\smash{\widetilde{\mathcal{A}}}}
\section{Introduction}
Despite the recent popularity of Quality-Diversity (QD) algorithms \cite{pugh2016quality, cully2017quality}, these algorithms have been limited to domains in which evaluations can be performed in simulation. This is because QD algorithms need to perform evaluations in the order of millions and where the outcomes are not safety critical or dangerous. Examples of these application domains include robotics \cite{cully2013behavioral, cully2015robots, chatzilygeroudis2018reset}, video games \cite{gravina2019procedural, fontaine2020illuminating} and aerodynamics \cite{gaier2018data}. In the field of robotics, physics simulators \cite{coumans2020, Lee2018, todorov2012mujoco} are commonly used and QD algorithms depend heavily on these to obtain abundant amounts of data and evaluations to learn behavioural repertoires of robots. However, building fast and accurate physics simulators to model the complex dynamics of robots and the wide variety of potential environments is difficult. Furthermore, even with extensive modelling of different scenarios, there is still the difficult problem of sim-to-real transfer \cite{zhao2020sim, akkaya2019solving, lee2020learning}. 
To realize the potential of learning and QD algorithms for robotics and to have the real-world impact we want them to have, we need algorithms which can effectively learn repertoires of skills autonomously and adapt directly in the real-world. 
%

\begin{figure}
\centering
    \includegraphics[width=0.45\textwidth]{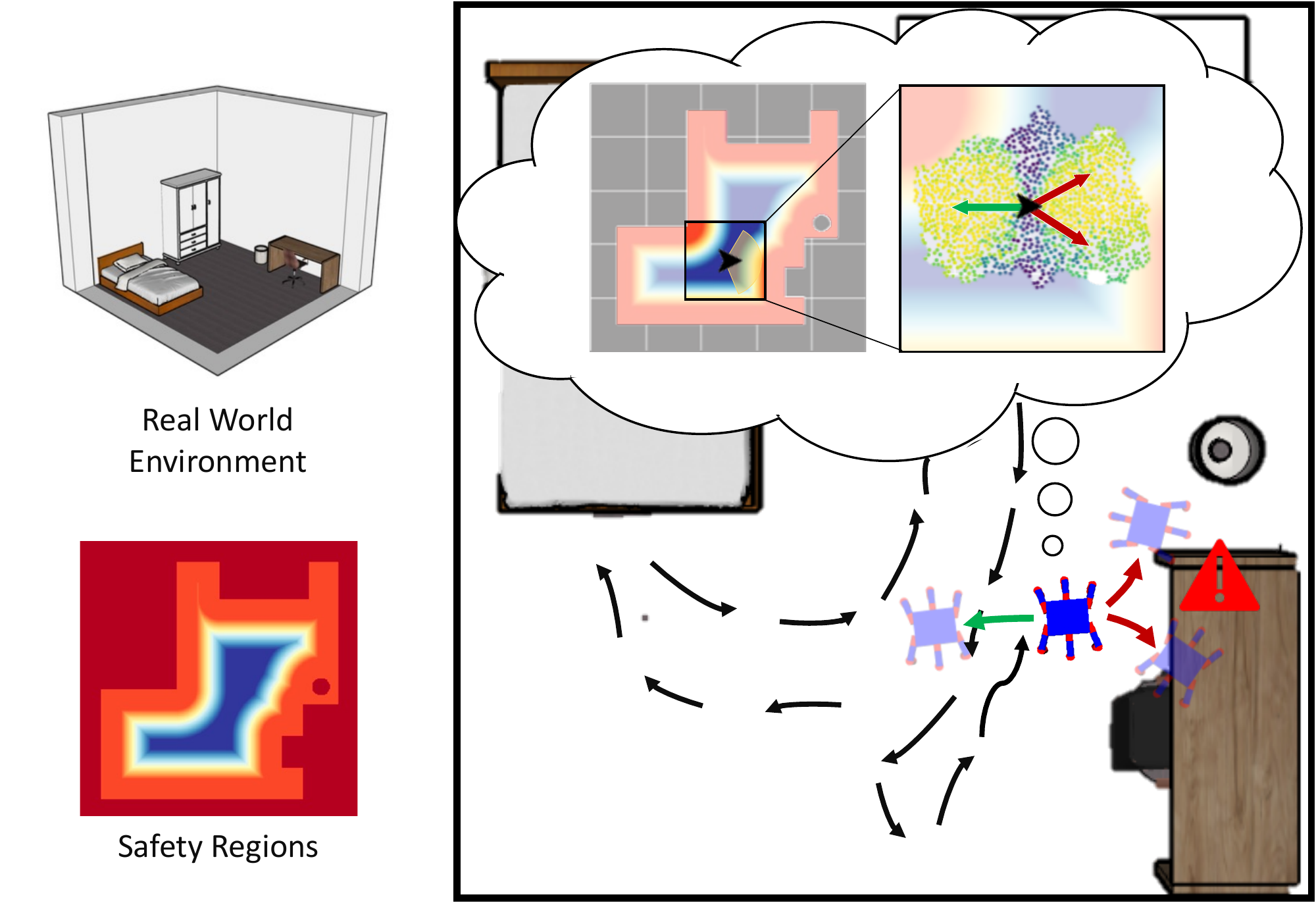}
    \caption{Safe reset-free movement through diversity of behavioural space demonstrated in a real-world environment.}
    \label{fig:illustration}
    \vspace{-6mm}
\end{figure}

In this work, as a step towards using QD algorithms in the real-world, we address two key issues that arise when attempting to learn behavioural repertoires in the real-world in an autonomous manner: \textit{resets} and \textit{safety}. 
We then aim to maximise the sample efficiency of learning behavioural repertoires while considering these constraints. We specifically focus on reset-free learning of robotic locomotion skills to highlight these issues and our approach to solving them. 

An often overlooked requirement of QD algorithms when used for Reinforcement Learning (RL) is the episodic setting they function in. This requires the environment to be set to a fixed initial state at the start of every episodic trial as the behavioural descriptor is measured as a function of the trajectory of states from this initial state to the final state in the episode. In real-world settings, this would correspond to humans manually resetting the robot and environment after every episode. This is an impractical and expensive solution considering the number of evaluations that conventional QD algorithms require are in the order of millions. With the considerable amount of human supervision and instrumentation required for resets, this defeats the purpose of QD algorithms to autonomously learn complex skills. 

Another key challenge of learning in the real-world is safety. Actions taken must not be dangerous to the robot and the environment. For learning locomotion skills, this corresponds to avoiding collisions with objects in the environment during learning. Achieving this while performing QD would require both the capability to predict the outcome of the execution of a new behaviour as well as an understanding of its implications with respect to its safety.

Finally, we want to learn skills efficiently with efficiency being measured by the number of evaluations taken. In the real-world, this also directly corresponds to the learning time needed. The goal is to intelligently select behaviours that would help improve the behavioural repertoire while minimizing unnecessary non-informative trials.

We introduce Reset-Free Quality Diversity (RF-QD) as a framework for the real-world execution of QD algorithms (see Figure~\ref{fig:illustration}). In a nutshell, RF-QD is a Dynamics-Aware Quality-Diversity (DA-QD) algorithm combined with a Behaviour Selection Policy to select only safe and valuable behaviours for evaluation in the (potentially dangerous) real-world. 
We demonstrate an algorithm which autonomously acquires a diverse repertoire of locomotion skills on a hexapod robot in safety-constrained environments.


\section{Related Work}

\subsection{Quality-Diversity and Behavioural Repertoire Learning in Robotics} \label{subsec:rel-QD}
Quality-Diversity (QD) optimization is a class of algorithms that aims to generate a collection of both diverse and high-performing solutions \cite{pugh2016quality, cully2017quality}. In the context of robotics, each solution can for instance, be a parametric policy which determines sequences of actions to execute (i.e. motor commands), resulting in a behaviour. A behaviour can then be represented by a numerical vector referred to as the \textit{Behavioural Descriptor (BD)}. The BD is a low-dimensional representation of the trajectory of states the policy visited and is usually defined manually depending on the tasks. However, the BD can also be learned in an unsupervised manner \cite{cully2019autonomous, paolo2020unsupervised, grillotti2021unsupervised}. The choice of the BD is important as it determines the novelty of a solution and will be used to drive the exploration to cover the BD space \cite{lehman2011abandoning}. 

Conceptually, QD extends the single novelty-seeking objective introduced in the Novelty Search algorithm \cite{lehman2011abandoning} with another measure of quality. MAP-Elites \cite{mouret2015illuminating} and Novelty Search with Local Competition \cite{lehman2011evolving} are two commonly used and well-known QD algorithms. Cully and Demiris \cite{cully2017quality} suggested that most QD algorithms can be represented using a common framework consisting of two key components; the \textit{archive} and \textit{selector}. Variants of QD algorithms develop around these components, all building on the QD loop of selection, variation, evaluation and (tentative) addition to the archive. The archive is used to store the highest performing solutions for each niche. Instead of a uniform grid used to discretize the BD space, methods like CVT-MAP Elites \cite{vassiliades2017using} or Sliding-Boundary MAP-Elites \cite{fontaine2019mapping} modify this to make the archive more flexible. More recently, there has also been a body of work focused on using more complex selectors and efficient optimizers such as evolutionary strategies \cite{colas2020scaling, fontaine2020covariance} and policy gradients \cite{nilsson2021policy, pierrot2021diversity}.

In the field of robotics, it was shown that the final archive of solutions discovered by QD algorithms can be used as a behavioural repertoire to perform downstream tasks \cite{cully2013behavioral}. For example, each solution is a controller which is represented by a set of parametric functions which controls the robot's joints. Coupled with Bayesian optimization, the controllers generated in the archive can be used to quickly adapt to unforeseen mechanical damage, help with sim-to-real transfer and solve longer-horizon tasks using planning algorithms \cite{cully2015robots, chatzilygeroudis2018reset}.
\begin{figure*}
\centering
    \includegraphics[width=0.85\textwidth]{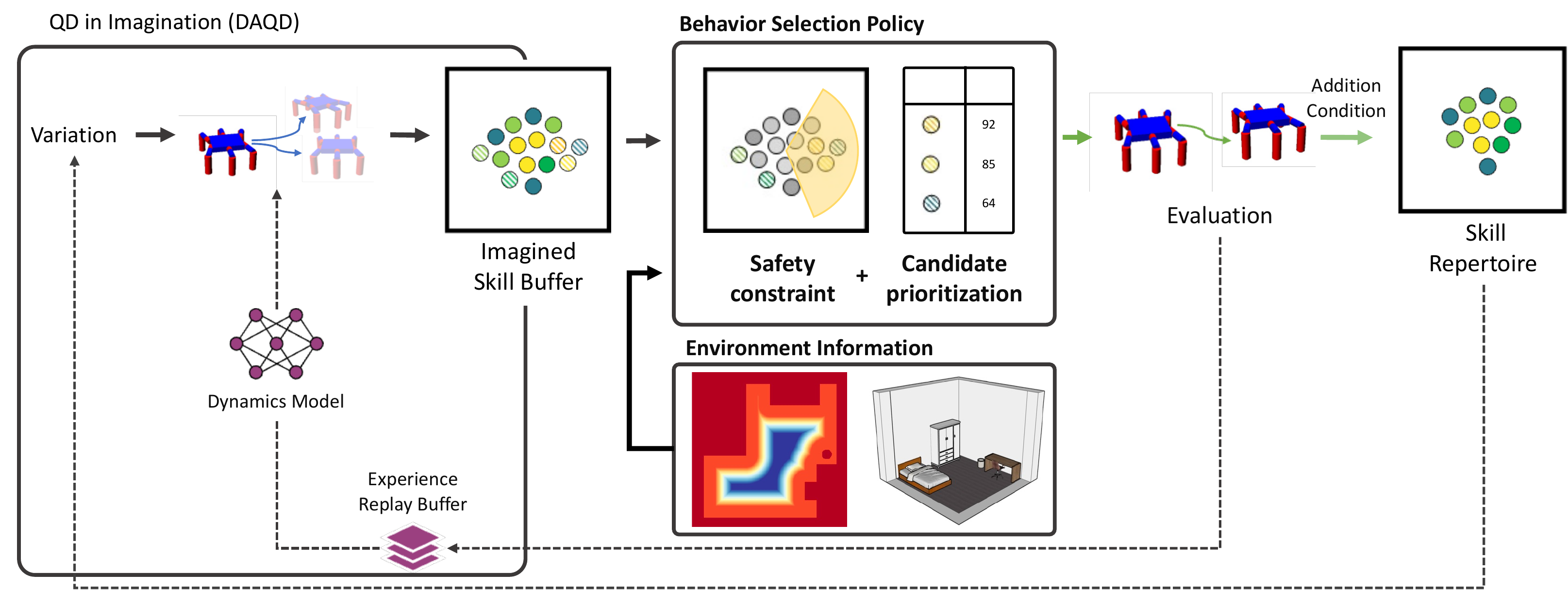}
    \caption{RF-QD performs QD in imagination (as in DA-QD) and uses a more intelligent behaviour selection policy to keep the robot in the safe regions of its environment while maximising the value gained by every real-world evaluation.}
    \label{fig:overview}
    \vspace{-4mm}
\end{figure*}
\subsection{Model-based Quality-Diversity} \label{subsec:rel-MBQD}
One of the main bottlenecks of Quality-Diversity (QD) algorithms is the sample efficiency. QD algorithms typically require on the order of millions of evaluations and rely on parallel computation of these evaluations. This single factor alone usually make them unsuitable to be used directly in the real-world.

A line of work that attempts to address this problem, now referred to as \textit{model-based quality-diversity} methods, is through the use of models. Surrogate-Assisted Illumination (SAIL) \cite{gaier2018data} first introduced the use of surrogate models for QD algorithms. SAIL integrates surrogate models, in the form of Gaussian Process (GP) models, to approximate the objective function and reduce the number of evaluations for the computationally expensive application of aerodynamic design. Another algorithm called M-QD \cite{keller2020model} later follows up on this idea and used neural network models that map the parameter space to the behaviour and fitness space as a surrogate model. They demonstrate this on robotic pushing and placing tasks.

Dynamics-Aware Quality-Diversity (DA-QD) \cite{lim2021dynamics} is another approach which instead uses learnt dynamics models as a surrogate model. DA-QD introduced the concept of the imagined repertoire which allows QD to be performed fully in imagination using the learnt dynamics models. The dynamics models are trained incrementally and online as data is collected through evaluations. DA-QD showed a significant ($\approx$20 time) increase in sample-efficiency. Our work uses the DA-QD framework to heavily reduce the number of evaluations needed making it feasible to be considered for a real-world application. With RF-QD, we extend DA-QD to make better use of the imagined repertoire to select behaviours more intelligently and in a sequential manner. While DA-QD only uses variation operators for optimization in imagination, we also further study the effect of optimizing for different objectives in imagination.

\subsection{Reset-free Learning} \label{subsec:rel-resetfree}
Reset-free Learning has mainly been studied in gradient-based Deep Reinforcement Learning (DRL) where the episodic setting is also usually a prerequisite of the Markov Decision Process (MDP) formulation of the problem. One approach taken to enable real-world RL is to automate resets using other manually scripted robots to reset objects and the environment to the initial state distribution required \cite{nagabandi2020deep}. While this works well for resetting manipulation tasks in which the workspace is relatively limited, this approach is difficult to apply for learning locomotion behaviours.

Most similar to our work and another promising method is to use a multi-task RL approach \cite{ha2020learning, gupta2021reset}. The key idea behind this approach is to use a scheduler and the different tasks present in the multi-task setup as resets for each other. Ha et al. \cite{ha2020learning} showed this in the context of learning simple locomotion policies while Gupta et al. \cite{gupta2021reset} demonstrated this approach on more extensive multi-task setting to learn dexterous manipulation policies. Both these works explicitly learn policies for tasks in a pre-defined distribution of tasks. Each policy is optimized individually using an off-the-shelf deep RL algorithm and separate instances of networks and replay buffers. Our work instead concurrently learns a repertoire of diverse policies using QD algorithms and leverage the diversity of the behavioural repertoire as resets. In Multi-task MAP-Elites \cite{mouret2020quality}, it is also showed that the behaviour space can also be viewed and formulated as a task-space where each cell is a task.

In the context of QD algorithms and behavioural repertoire learning, the Reset-Free Trial and Error (RTE) \cite{chatzilygeroudis2018reset} algorithm has also aimed to address the reset problem. RTE demonstrated this for adaptation using the behavioural repertoire as a prior for Gaussian Process models. This is a different setting from the work we present in this paper as the behavioural repertoire generation process itself in RTE is performed fully in simulation using resets. The reset-free in RTE refers to the reset-free adaptation when performing sim-to-real transfer or reset-free adaptation to mechanical damage. In our work, we aim to learn the behavioural repertoire itself in a reset-free manner.
\section{Background: Dynamics-Aware QD} \label{sec:background}
We build on the DA-QD framework proposed by Lim et al. \cite{lim2021dynamics}. We briefly summarize DA-QD here and refer the reader to the full paper \cite{lim2021dynamics} for further details. DA-QD is a model-based QD algorithm which extends the conventional QD framework \cite{pugh2016quality, cully2017quality} discussed in section \ref{subsec:rel-QD} with three key components: a \textit{dynamics model}, an \textit{imagined repertoire} and \textit{selector} from the imagined repertoire.

The learnt dynamics model is a forward dynamics model and is represented by a neural network parameterized by $\theta$. To capture both aleatoric and epistemic uncertainties, an ensemble of probabilistic models are used.
This forward model can be represented as $\dynamicsmodel(\vec s_{t+1} | \vec s_t, \vec a_t)$.

Here, the disagreement between predictions of all models in the ensemble captures the epistemic uncertainty, i.e. it indicates the uncertainty of the prediction due to a lack of samples. The overall model disagreement $\mu_d$ can be calculated as the expected difference between any two models in the ensemble $f_\phi$ for one state-action pair, averaged over all time step predictions in one rolled-out trajectory (i.e. one evaluated behaviour) of length T~\cite{Kidambi2020}:
\begin{equation} \label{eqn:disagreement}
    \begin{split}
        \text{disag}(s,a) = \mathop{\mathbb{E}}_{i \neq j} \| f_{\phi_i}(s,a) - f_{\phi_j}(s,a) \|_2 \\
        \textstyle \mu_d = \frac{1}{T} \sum_{t=0}^T \text{disag}(s_t,a_t)
    \end{split} 
\end{equation}
State transition data is collected and stored in a replay buffer $\replaybuffer$ as evaluations of robot behaviour are performed in the environment. The model is trained in a self-supervised manner to maximise log-likelihood of the transitions sampled from replay buffer and is optimized via back-propagation.

The dynamics model $\dynamicsmodel$ can be called recursively to evaluate policies in what is referred to as an imagined roll-out. The expected fitness and BD can be obtained from this imagined roll-out as both these quantities measured are a function of the state trajectory. DA-QD introduced the concept of an imagined repertoire $\archivesynthetic$ to organise and maintain solutions that have been evaluated in imagination using the dynamics model. The imagined repertoire $\archivesynthetic$ uses the the same addition conditions as the repertoire $\archivereal$. The imagined repertoire $\archivesynthetic$ only allows solutions that have been evaluated in imagination that are expected to be novel or better performing than existing solutions to be considered for evaluation. This is where the sample-efficiency of this method is derived from. Additionally, this allows QD to be performed fully in imagination. This means that the selection and mutation of the QD algorithm can be continuously performed from the imagined archive for any desired number of imagined generations without any samples or evaluations on the real system.

Finally, with the introduction of the imagined repertoire $\archivesynthetic$, this necessitates selection of solutions from $\archivesynthetic$ to be evaluated. As the original DA-QD algorithm does not consider the reset-free sequential evaluation setting, the authors select all the solutions that have been added to the imagined archive to be evaluated in parallel. Our work extends DA-QD and proposes a more intelligent method to select and manage solutions in the imagined archive given the reset-free setting and safety constraints from the environment.
Additionally, DA-QD also does not explicitly use the resulting model-disagreement. In this paper, the model-disagreement is used both as a heuristic to select behaviours to execute more intelligently (Sec.~\ref{subsec:methods-behaviour-selection}) and as an optimization objective in imagination (Sec. \ref{sec:results_emitters}). 

\section{Methods}

We present Reset-Free Quality-Diversity (RF-QD) as a method to enable the application of QD's behavioral repertoire learning in non-episodic real-world environments (see Algorithm \ref{Algo:RF-QD_short}). We  treat the robot as an actor in its environment that performs a constant search for new and improved behaviours and storing these in the archive. For this, we extend the classical QD loop by two steps. Firstly, we build on the pre-evaluation of any new behaviour "in imagination" by a dynamics model (DA-QD). Secondly, we introduce a behaviour selection policy, that modulates the robot's search for novel and high-performing behaviours as to comply with the safety constraints given by the environment (see Figure \ref{fig:overview}).

In the following, we first elaborate on the core of our method: the \textit{behaviour selection policy}. Then, we detail its main components: the \textit{safety evaluation}, \textit{safety constraints}, the \textit{prioritisation metrics} and the \textit{recovery policy}.
\begin{algorithm}[H]
\begin{algorithmic}
\State \textbf{Input:} archive $\mathcal{A} = \emptyset$, candidate buffer $\mathcal{C} = \emptyset$,\\ dynamics model $\dynamicsmodel$, experience replay buffer $\mathcal{B}$

\State
\State $\mathcal{A}, \mathcal{B} \leftarrow$ random\_archive\_initialization()
\State $\dynamicsmodel \leftarrow$ train\_model($\mathcal{B}$)
\State
\For{max\_iterations}
    \If {$\mathcal{C} == \emptyset$}
    \Comment Fill the candidate buffer
        \State $\archivesynthetic \leftarrow$ generate\_candidates($\mathcal{A}, \dynamicsmodel$)
        \Comment Using imagination
        \State $\mathcal{C} \leftarrow \archivesynthetic \backslash \mathcal{A}$
    \EndIf
    \State $s \leftarrow$ get\_robot\_state()
        \Comment Get state to evaluate safety
    \If{$s$ is safe}
    \Comment Apply candidate selection policy
        \State $\mathcal{C}_{safe} \leftarrow$ apply\_safety\_constraint($\mathcal{C}$)
        \State x $\leftarrow$ argmax(prioritize\_candidates($\mathcal{C}_{safe}$))
        \State $b_x, f_x \leftarrow$ execute\_behaviour(x)
        \State $\mathcal{A} \leftarrow$ add\_to\_archive(x, $b_x$, $f_x$)
    \Else
    \Comment Apply recovery policy to return to safety
        \State x $\leftarrow$ recovery\_policy($\mathcal{A}$)
        \State execute\_behaviour(x)
    \EndIf
    $\mathcal{B} \leftarrow$ add\_to\_replay\_buffer()
    \State $\dynamicsmodel \leftarrow$ train\_model($\mathcal{B}$)
\EndFor

\end{algorithmic}
\caption{Reset-free Quality-Diversity (RF-QD)}
\label{Algo:RF-QD_short}
\end{algorithm}

\subsection{Reset-free Behaviour Selection} \label{subsec:methods-behaviour-selection}
To be able to stay safe while acting in its environment, we introduce a behaviour selection policy to modulate the robot's actions in the real world. This behaviour selection ensures that every new behaviour will only be performed if it is expected to be safe for the robot. 
At every step, new behaviours are selected from a candidate buffer $\mathcal{C}$. The candidate buffer
$\mathcal{C}$ is regularly filled with new policies from the imagined repertoire $\archivesynthetic$ that are not already present in the repertoire $\archivereal$. 
$\archivesynthetic$ is a component introduced in DA-QD that maintains solutions that were evaluated in imagination using the dynamics model $\dynamicsmodel$.
Based on the robot's current state in the environment, our policy then selects a subset of candidate behaviours $\mathcal{C}_{safe}$ that have a low risk of violating the safety constraints given by the environment. Out of these, the candidate behaviour that has the highest projected prioritization score will then be evaluated in the real world. In the following sections, the core components will be described in more detail.

\subsection{Safety Evaluation}
In this paper, we assume knowledge of the environment layout, represented by 'safety regions' (see Figure \ref{fig:illustration}), that indicate the region of dangerous states $\Omega$. In practice, this information could as well be obtained using Simultaneous Localisation and Mapping (SLAM) methods with an on-board camera. 

Derived from the robot's state $s$, we define the exploration parameter $\epsilon(s)$, which indicates the relative degree of safety in the current state. It is calculated as the smallest distance between $s$ and $\Omega$ and normalised by the maximum encountered distance value (see Equation \ref{eqn:epsilon_general}). While inside the safe region (i.e. $s \notin \Omega \rightarrow \epsilon(s) > 0$), the robot must choose any potential solution to be evaluated in the real world that is predicted to keep $\epsilon(s) > 0$, i.e. does not enter any unsafe state.
To lower the risk of damage to the robot, an offset $\beta$ can be added in the computation of $\epsilon_s$ as an increased threshold for the minimum distance towards the border of the region of unsafe states within the state space.

\begin{equation}
\label{eqn:epsilon_general}
    \epsilon(s) = \frac{dist(s, \Omega) - \beta}
                {\underset{s_i}{\max} \, dist(s_i, \Omega) - \beta}
\end{equation}

From the dynamics model, we can obtain the predicted next state $s'$ after the execution of a candidate behaviour and compute $\epsilon(s')$. $s'$ corresponds to the state $s_{t+T}$ after $T$ timesteps, where $T$ is the length of one behaviour.
Generally, we seek the robot to stay as close as possible to the safest point(s) in the environment, i.e. maintain maximal distance to the region of dangerous states ($\epsilon(s) \approx 1$).

\subsection{Safety Constraints}
\begin{figure}
\centering
    \includegraphics[height=0.15\textwidth]{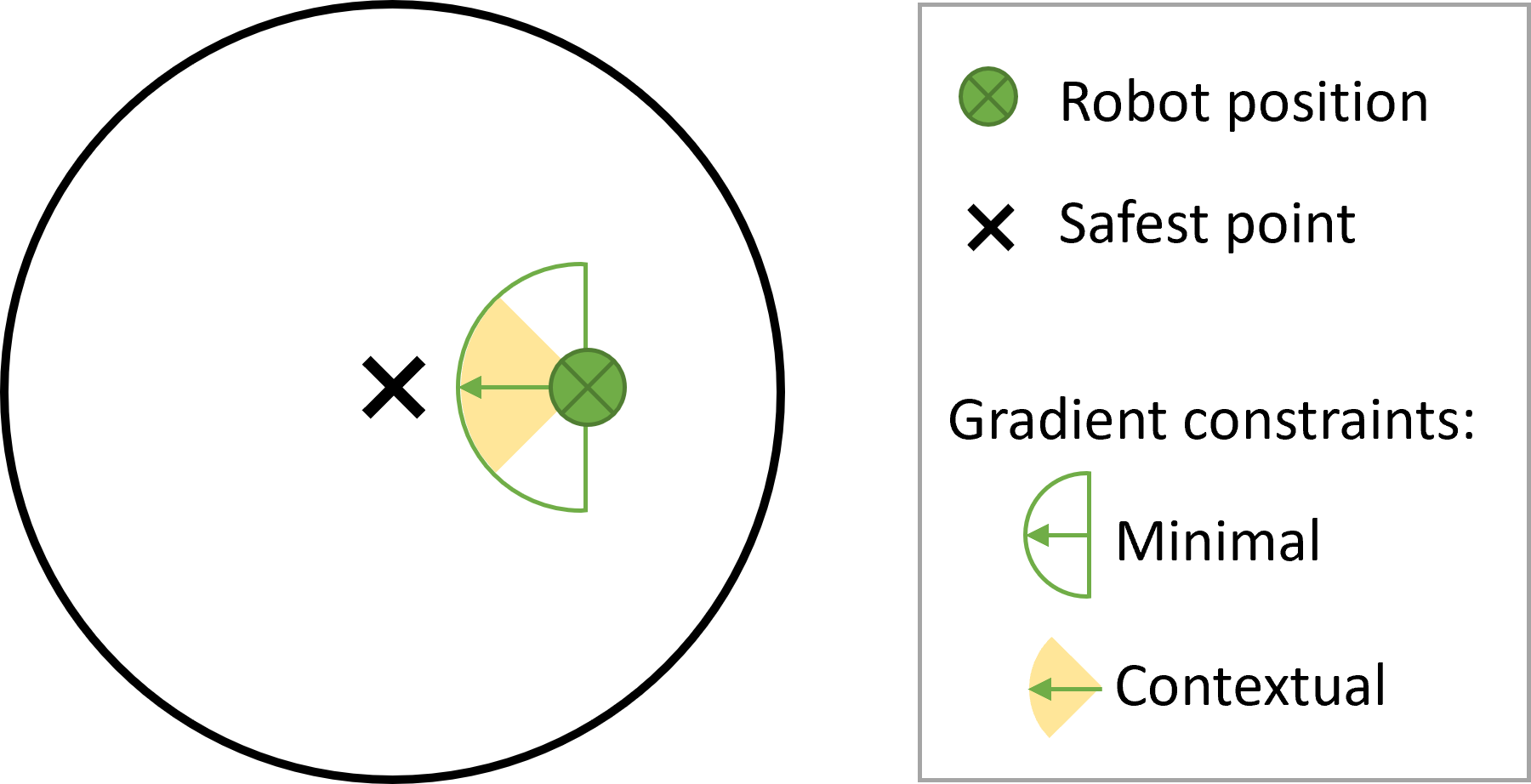}
    \caption{Sketch of the gradient-based safety constraints in a simple circular 2D-environment.}
    \label{fig:safety_constraint}
    \vspace{-4mm}
\end{figure}

For every behaviour selection performed by our policy, we first employ a safety constraint to determine the safe subset $\mathcal{C}_{safe}$ of all available candidate behaviours with respect to the robot's current state. We can use different constraints depending on our knowledge of the environment and the intended risk aversion of our exploration. In the experiment section below, we evaluate the following constraints, all of which are based on the predicted robot state $s'$ after the execution of each imagined behaviour (given the current robot state $s$):
\begin{itemize}[leftmargin=*]
\setlength\itemsep{1em}
    \item As a \textit{minimal constraint} we consider only candidate behaviours with $\epsilon(s') > 0$ to ensure we never execute a behaviour that was already expected to be unsafe. 

    \item Alternatively, a \textit{contextual constraint} carries weight only if the current robot state is near the border of the region of unsafe states ($\epsilon(s) \approx 0$), but enables free exploration if it is far away from potential danger ($\epsilon(s) \approx 1$):
    \begin{equation}
    \label{eqn:constraint_minimal}
        \epsilon(s') > \epsilon(s) \cdot (1-\epsilon(s))
    \end{equation}
    
    \item If we have access to the gradient of the epsilon function, the direction of maximal improvement of safety with respect to the next state can be computed as $\nabla_s \epsilon(s)$. The \textit{gradient-minimal constraint} considers only solutions moving in the general direction of the gradient. Based on the dot product of the unit vectors of the gradient of the epsilon function ($\nabla_s \epsilon(s)$) and the projected movement in state space ($s'-s$), we formulate a lower bound for deviation from the direction of the gradient as:
    \begin{equation}
    \label{eqn:constraint_gradient_minimal}
        \frac{s'-s}{||s'-s||}\cdot\frac{\nabla_s \epsilon(s)}{||\nabla_s \epsilon(s)||} \geq 0
    \end{equation}
    Geometrically, this is equal to a maximum deviation of 90° in 2D space as visualised in Figure \ref{fig:safety_constraint} (green semicircle).
    
    \item Again, we can modify this into a more strict \textit{gradient-contextual constraint} by using the value of epsilon at the current state of the robot to modulate the constraint. This way, the constraint is more relaxed towards the centre of the region of safe states but only accepts small deviations from the direction of the safety gradient close to the border of the region of unsafe states:
    \begin{equation}
    \label{eqn:constraint_gradient_contextual}
        \frac{s'-s}{||s'-s||}\cdot\frac{\nabla_s \epsilon(s)}{||\nabla_s \epsilon(s)||}
        \geq \epsilon(s) \cdot (1-\epsilon(s))
    \end{equation}
    Geometrically, this is equal to a deviation from the gradient proportional to $\epsilon(s)$ (see yellow region in Figure \ref{fig:safety_constraint} for $\epsilon(s)=0.5$).
    
    \item Finally, safety can also be enforced not by a hard constraint, but as a component of the prioritization measures. This can especially be useful as a supplement to the gradient-free constraints in complex environments.
\end{itemize}

\subsection{Prioritization Metrics}
After the safe subset of candidate behaviours $\mathcal{C}_{safe}$ has been selected based on the safety constraint, the remaining candidates are ranked according to a prioritization measure as the second step in behaviour selection. This is intended to give priority to the real-world evaluation of candidate behaviours which have the highest value for the overall QD algorithm performance, as real-world samples are expensive to collect. 
Finally, the candidate with the highest prioritization score is selected. The composition of prioritization measures can be adapted depending on the task at hand. We can either use a single prioritization measure or a (weighed) sum of multiple values. In this work, we have evaluated the following measures:
\begin{itemize}[leftmargin=*]
\setlength\itemsep{1em}
    \item Firstly, the robot's \textbf{safety} can be considered again as a prioritization measure through the dynamic exploration parameter $\epsilon(s')$ as outlined above. Generally, this approach will be used in combination with another metric to enable the behaviour policy to tolerate a possible safety violation in favor of a higher score.
    
    \item Another key measure to score a candidate behaviour is the \textbf{dynamics model disagreement}. 
    The dynamics model used in DA-QD consists of an ensemble of models to capture the epistemic uncertainty via disagreement between the predictions of the models (see Section \ref{sec:background} and Equation \ref{eqn:disagreement}). 
    The epistemic uncertainty can also be interpreted and formalised as an information theoretic measure of the expected information gain \cite{pathak2019self, sekar2020planning}. 
    Maximising the model-disagreement has been used as a self-supervised intrinsic reward for exploration in Deep RL literature \cite{pathak2019self, sekar2020planning}. 
    The key idea behind this measure is to prioritise policies that are most informative based on our current knowledge which is represented via the ensemble of dynamics models (i.e. epistemic uncertainty).
    Selecting policies with high-model disagreement would mean visiting states that have been less explored than others.
    As we incrementally train the dynamics model on incoming data, policies that visit states that have been seen will no longer have a large model disagreement which will allow this measure to continuously be used to explore.
    Depending on the state of the robot in the environment, we can prioritize high and low model disagreement behaviours. 
    Conversely, policies with low disagreement should be prioritized in safety-critical situations. Solutions with low expected model disagreement are likely to resemble the expected outcome and indicates the model's confidence.
    
    \item Finally, we also consider the classical metrics used to quantify behaviours in QD. This is firstly the \textbf{novelty} of a candidate behaviour as the distance to the k nearest solutions already in the archive ($\nu_1, ..., \nu_k$)~\cite{lehman2011abandoning}. Similarly, we could also consider the quality of a solution through a measure such as the QD improvement~\cite{fontaine2020covariance} or the future value of a solution through its curiosity score~\cite{cully2017quality}. However, this is left for future work.
\end{itemize}

\subsection{Recovery Policy}
As a final safeguard to keep the robot in the safe region of the environment, we introduce a recovery policy to return the robot to safety if it ever violates any of the environment's safety constraints. These constraints can be derived from the environment in various ways, e.g. as a minimum distance to obstacles represented by 'safety regions' as in this work. Should the robot leave the safe region, the discovery of new behaviours will be halted and a greedy behaviour selection policy will be employed over the archive of behaviours that were already evaluated in the environment instead of the buffer of candidate behaviours. Here, we pick the single behaviour that is projected to effect the greatest improvement in safety.

\section{Experiments}
We evaluate our method with an 18 DoF hexapod robot on an adapted version of the omni-directional locomotion task~\cite{cully2013behavioral}.
In this task, the robot learns behaviours to walk in every direction from an initial position.
For the controllers, we evolve parameters of a sinusoidal control signal that is sent to each motor. This sinusoidal signal acts as a structural prior towards periodic movement for locomotion.
As we focus on a reset-free setting, all evaluations of new behaviours have to be done sequentially and cannot be parallelised.
All simulations are performed in RobotDART building on the Dynamics Animation and Robotics Tookit (DART) simulator \cite{dartsim}. 
To simulate a practical number of trials that would be performed in the real-world experiment, the number of evaluations performed in any single run of the algorithm are limited to 10,000.

\subsection{Baseline comparison}
\label{sec:results_baselines}
\begin{figure}
\centering
    \includegraphics[width=0.45\textwidth]{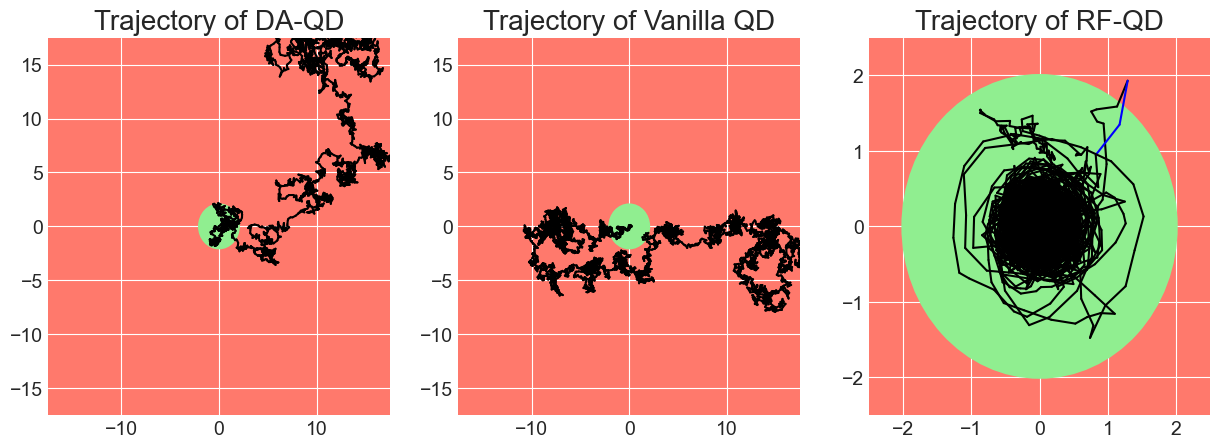}
    \caption{Example trajectories of DA-QD, Vanilla-QD and RF-QD in flat environment with safe region (green) and dangerous region (red).}
    \label{fig:trajectory}
    \vspace{-4mm}
\end{figure}

Firstly, we evaluate the general capability of the RF-QD method.
For this, we compare against "vanilla" QD and DA-QD \cite{lim2021dynamics} as baselines. RF-QD and both baselines use the Iso-dd~\cite{vassiliades2018discovering} variation operator.
We use a simple flat environment with a circular region of safety with radius $r=2.0m$.
Figure \ref{fig:trajectory} shows example trajectories of the baselines compared to RF-QD. The baselines' random selection of behaviours causes the robot to trail off deeply into the dangerous region, while RF-QD performs its exploration almost entirely within the safe region. The depicted RF-QD run leaves the safe region once, but then deploys the recovery policy (blue line) to return to safety.

As the baseline methods are not made for a reset-free environment, for all further comparisons we perform manual resets to the starting position if the robot leaves the safe region by more than 50 cm. This is similar to what is done when performing QD on a real-world robot today. For the baseline comparisons, RF-QD was run with a gradient-contextual safety constraint and encouraging maximal novelty through the prioritization strategy. This configuration has proven powerful in our evaluation of different constraints and prioritization measures. Table \ref{tab:baselines_safety} quantifies the safety of the three algorithms averaged over 10 replications of each. We can see, that RF-QD achieves almost perfect safety - never once requiring a safety reset as described above and only rarely taking a single step outside the safe region.

\begin{table}
  \caption{Safety metrics for all variants, averaged over 10 runs (mean ± std).}
  \label{tab:baselines_safety}
  \begin{tabular}{c|ccc}
    \toprule
    Variant     &Resets &Steps outside safety  &Recovery steps\\
    \midrule
    Vanilla-QD  & 54.0 ± 4.2    & 908.0 ± 74.1    & n/a  \\
    DA-QD       & 114.0 ± 17.8  & 1039.5 ± 51.0   & n/a  \\
    RF-QD       & 0.0 ± 0.0     & 1.0 ± 2.8       & 3.5 ± 9.9 \\
  \bottomrule
\end{tabular}
\end{table}

\begin{figure}
\centering
    \includegraphics[width=0.45\textwidth]{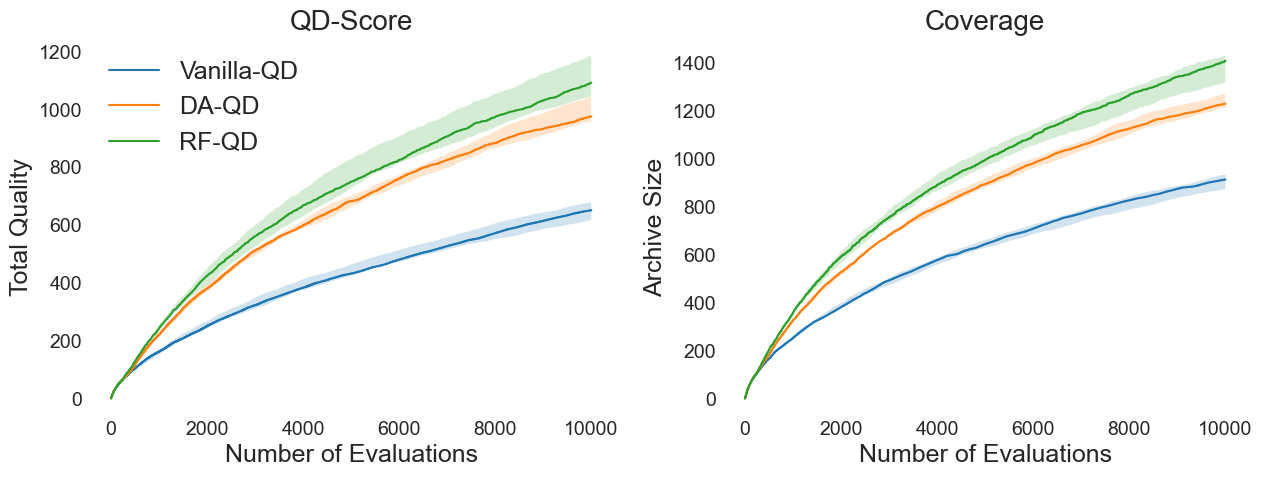}
    \caption{QD-Score and coverage of RF-QD and baselines on the circular safe area environment. The graphs represent the median as a coloured bold line, while the shaded area extends to the first and the third quartiles over 10 runs.}
    \label{fig:baselines_performance}
    \vspace{-4mm}
\end{figure}

Additionally, RF-QD slightly outperforms its direct baseline DA-QD in terms of both QD-score and coverage as shown in Figure \ref{fig:baselines_performance}. While the distance to vanilla QD is due to DA-QD's increased sample efficiency, RF-QD's behaviour selection policy does not sacrifice performance for safety, but even improves performance by its candidate prioritization strategy (i.e. novelty in this case).

\subsection{Comparison of Policy Configurations}
\label{sec:hyperparameters}
\begin{figure}
\centering
    \includegraphics[width=0.5\textwidth]{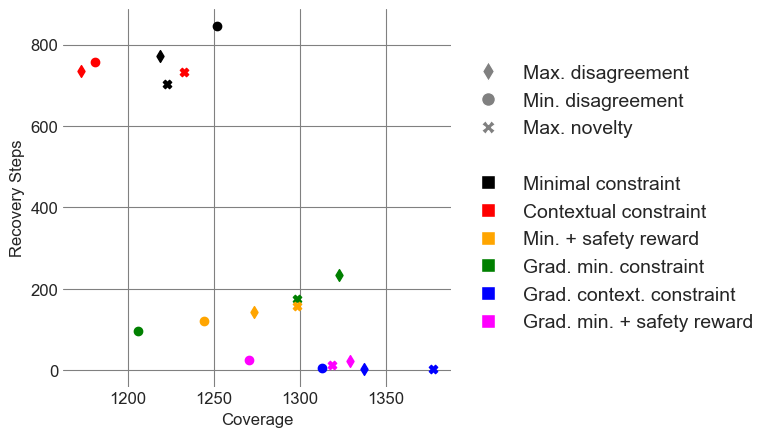}
    \caption{Comparison of different Behaviour Selection Policy configurations on both performance (coverage) and safety (recovery steps) on the circular safe area environment.}
    \label{fig:policy_comparison}
    \vspace{-4mm}
\end{figure}

Additionally, we evaluated the various configurations of the Behaviour Selection Policy as introduced in Section \ref{subsec:methods-behaviour-selection}. Figure \ref{fig:policy_comparison} shows an overview over the different combinations of safety constraints and prioritization measures. Here, the policy configurations are evaluated by performance (represented by their final coverage) and safety (represented by the number of recovery steps), both from runs of 10,000 steps over 10 replications. In short, Figure \ref{fig:policy_comparison} shows strong separation between the relatively unsafe minimal and contextual constraints (both gradient-free) and all remaining constraints. The strongest performance is exhibited by variants combining the novelty or disagreement maximising prioritization measures with a gradient contextual constraint. Out of the naive gradient-free constraints, which must be used if there is no single 'safest' direction of movement (as e.g. in more complex environments such as the one following in Section \ref{sec:results_complex}), only the soft constraints achieves comparable safety scores and performances as the gradient-based configurations. Which exact configuration should be chosen will however always depend on the exact task at hand.

\begin{figure*}
\centering
    \includegraphics[width=1\textwidth]{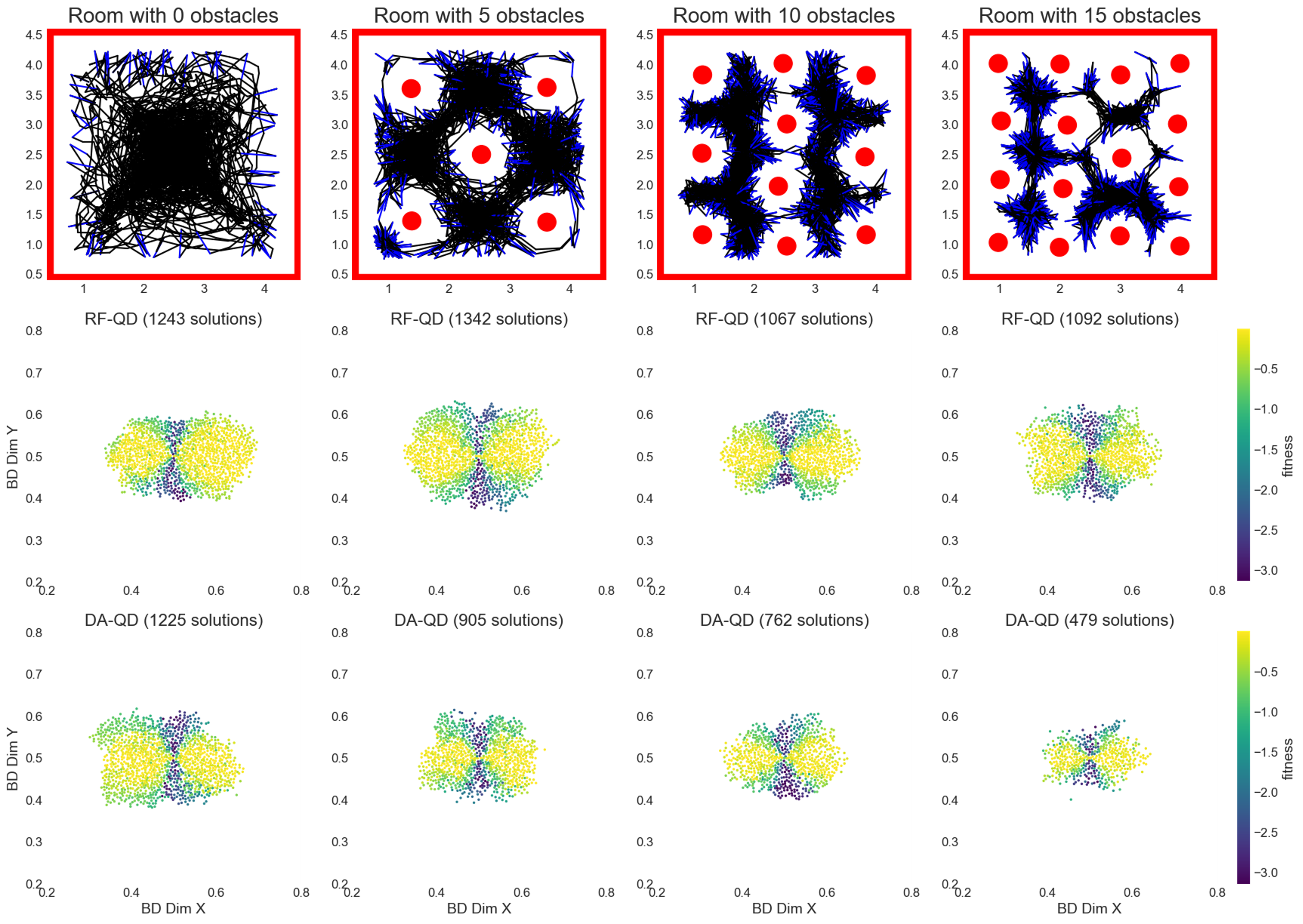}
    \caption{Complex environments with 0, 5, 10 and 15 obstacles. Top: Example trajectories of hexapod acting under RF-QD. Middle: Example archives by RF-QD. Bottom: Example archives by DA-QD.}
    \label{fig:environments}
    \vspace{-2mm}
\end{figure*}

\subsection{Robustness to environment complexity}
\label{sec:results_complex}
To evaluate RF-QD's performance in increasingly complex environments, we exchange the previous circular environment for a closed 4x4m room with a number of column-shaped obstacles. Figure \ref{fig:environments} shows examples of such environments including RF-QD's trajectories in them (top row). 
We can observe that the robot acting under RF-QD keeps its distance from the obstacles, while building archives of behaviours (middle row) that are radically less affected by the environment complexity than those created by DA-QD (bottom row).

\begin{figure}
\centering
    \includegraphics[width=0.5\textwidth]{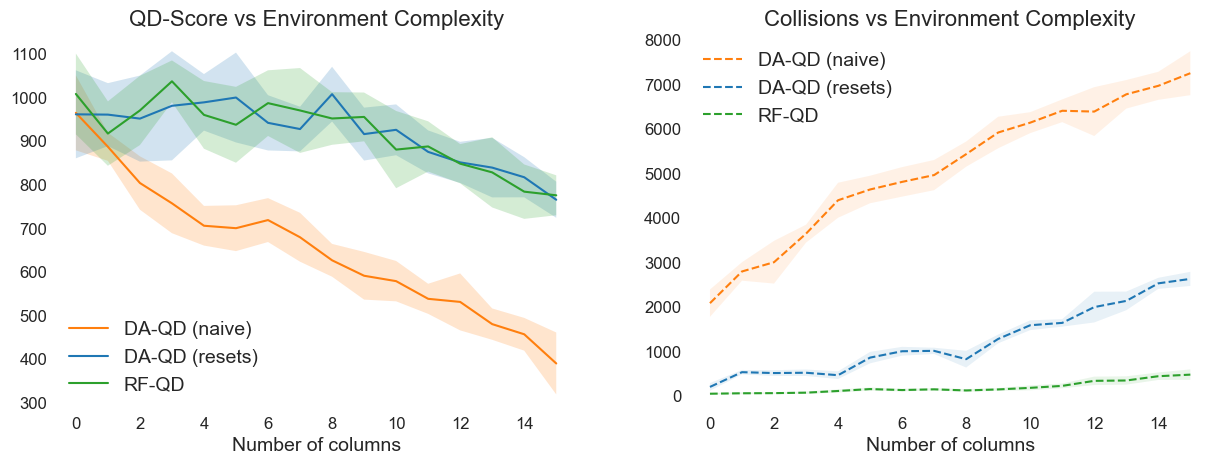}
    \caption{Increasingly complex environments: QD-Scores vs number of obstacles. The graphs represent the mean as a coloured bold line, while the shaded area extends to the standard deviations over 10 runs for each environment.}
    \label{fig:advanced_qd_scores}
    \vspace{-4mm}
\end{figure}

In these complex environments, we employed RF-QD with a safety-focused configuration. This uses a minimal (hard) safety constraint combined with two equally weighed prioritization measures to select behaviours that maximise safety (through $\epsilon$) and have low model disagreement. 
As a benchmark for QD performance, we again add a version of DA-QD that uses safety resets, now triggered on any collision with an obstacle. 
We also keep a 'naive' version of DA-QD, that is not reset upon collision (same as RF-QD). These algorithms were compared in rooms with 0 to 15 obstacles (see Figure \ref{fig:advanced_qd_scores}). 
While in an empty room, all algorithms perform similarly well, the naive DA-QD variant quickly drops in performance with a growing number of obstacles through a large number of collisions (which render the corresponding evaluations invalid). 
At the same time, RF-QD manages to fully keep up with the upper baseline of DA-QD (using safety resets). 
While a more performance-focused prioritization strategy (i.e. novelty as in Section \ref{sec:results_baselines}) for RF-QD might have increased QD-scores slightly, this would have sacrificed the safety of the robot in more challenging environments.

\subsection{Effect of objectives in imagination} \label{sec:results_emitters}

\begin{figure}
\centering
    \includegraphics[width=0.5\textwidth]{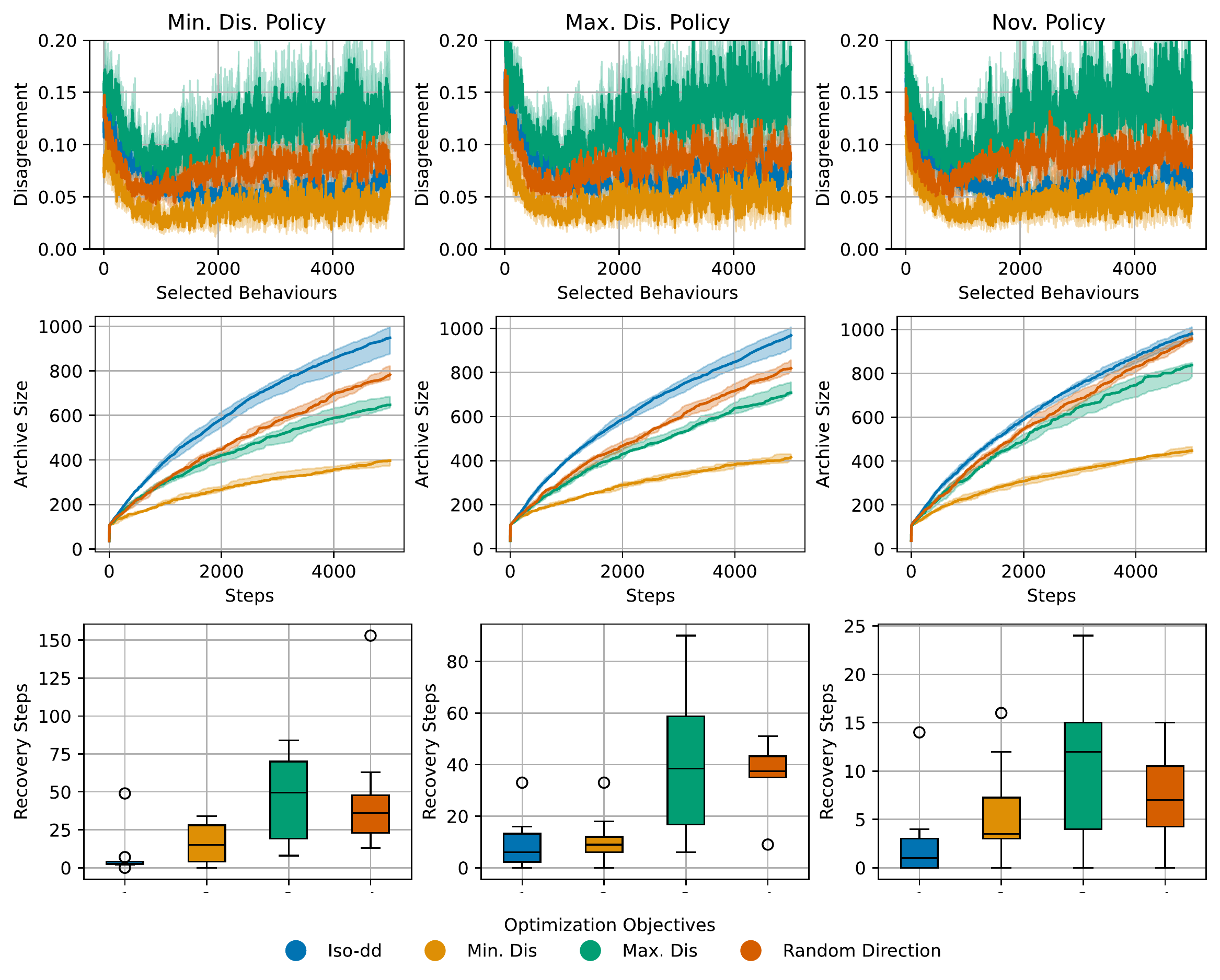}
    \caption{Study of different optimization objectives and prioritization metric configurations. Each panel considers a different prioritisation metric. Top: Disagreement of selected behaviours by RF-QD. The bold lines and shaded areas represent the median and interquartile range over 10 replications respectively. Middle: Progression of the archive size over the number of selected behaviours for each optimization objective. Bottom: Distribution of the total number of recovery steps for each optimization objective.}
    \label{fig:emitter_results}
    \vspace{-4mm}
\end{figure}
We also study the effect of the type of solutions available in the candidate buffer that the behaviour selection policy chooses from.
To study this, we investigate the influence of different optimisation objectives for the generation of the candidate buffer during the QD in imagination. 
When using Iso-DD \cite{vassiliades2018discovering}, the solutions are relatively generic and objective-agnostic, i.e., not optimised to fulfil a specific objective. 
Alternatively, we can use different types of emitters (introduced by CMA-ME \cite{fontaine2020covariance}) to produce solutions that maximise a specific objective.
We perform experiments using three different optimization objectives: maximising model disagreement, minimising model disagreement, and a random direction objective as a surrogate objective for novelty. We compare this to the standard Iso-dd variations used in all our experiments as a baseline.
We perform an ablation of these three different objectives with their corresponding prioritization measures used in the behaviour selection policy. We report results across 10 replications.

First, we evaluate the effect of more targeted objectives by analysing the model disagreement associated with the individuals selected by the behaviour selection policy (Figure \ref{fig:emitter_results}). 
The key take-away from Figure \ref{fig:emitter_results} (top) is that the optimisation objectives used when running QD in imagination can strongly influence the behaviours that are finally selected. 
We can see that regardless of the prioritization metric used by the behaviour selection policy, the same overall trends are always observed:
The minimising disagreement optimization objective (yellow) always results in low disagreement individuals being selected by the behaviour selection policy regardless of the prioritization metrics.
The same observation applied to the maximising disagreement objective (green).
This observation corresponds to our initial hypothesis where targeted optimization objectives can skew the distribution of solutions generated towards the target objective.
This results in a higher probability for the solutions with the desired metric being selected.

Given that biased/specialised sets of solutions can be generated in the candidate buffer using more targeted objectives, we evaluate the effect of the composition of this candidate buffer on the performance of RF-QD. 
Figure \ref{fig:emitter_results} (middle and bottom) show that the objective-agnostic Iso-DD operator outperforms all the targeted optimization objectives both in terms of coverage and safety (number of resets) across all prioritization measures used by the behaviour selection policy.
This is an interesting result as one could expect the variants with aligned prioritization measures and  optimization objectives to perform better.
We hypothesize that the buffer of candidate solutions being generated by targeted objectives become too specialised while the objective-agnostic Iso-DD can generate a diverse buffer of solutions to choose from.
This is not such a surprising observation as Multi-Emitter MAP-Elites \cite{cully2020multi} had previously also shown that when using simultaneously multiple emitter types, the random emitter (based on Iso-dd) remains the most fruitful through the entire process compared to other objective-driven emitters.

\section{Discussion} \label{label:discussion}
In this paper, we have presented RF-QD, a method to learn behavioural repertoires autonomously without resets in realistic environments. We demonstrate how an intelligent behaviour selection policy can be used with QD in imagination to learn safely and efficiently. We first test RF-QD to learn while remaining within a designated area and show that the behaviour selection policy is necessary to prevent the need for resets and to stay within the safe training area. 
We then show how RF-QD can also operate in more complex environments with many obstacles and minimal room for error.
Our results also show that we can acquire full repertoires despite increasing environment complexity while the performance of DA-QD and Vanilla QD baselines deteriorate with the increase in complexity. 
Lastly, we conduct an ablation to investigate the effect of the type of solutions present in the candidate buffer on the performance of RF-QD.
We demonstrate that using targeted optimization objectives when performing QD in imagination can bias the distribution of solutions presented to the behaviour selection policy. 
Our results show that it is important to keep the diverse types of solutions in the candidate buffer over just specialised solutions biased towards a single metric.

For future work, we also hope to show RF-QD learning directly on a real world system, with no dependence on simulators. Additionally, this paper only considers safety and danger in the form of obstacle avoidance. We leave other forms dangerous scenarios and work on safety detection for future work.
\begin{acks}
This work was supported by the Engineering and Physical Sciences Research Council (EPSRC) grant EP/V006673/1 project REcoVER. We would like to thank the members of the Adaptive and Intelligent Robotics Lab for their very valuable comments. 
\end{acks}

\bibliographystyle{ACM-Reference-Format}
\bibliography{biblio}

\appendix

\end{document}